\documentclass[sigconf]{acmart}

\usepackage{multirow}
\usepackage{subcaption}
\usepackage{enumitem}
\usepackage{cleveref}
\usepackage{colortbl}
\usepackage{makecell}
\usepackage{balance}

\newcommand{\nummodels}{194}

 \newcommand{\rhoVal}[1]{$\rho{=}#1$}

\AtBeginDocument{%
  }

\renewcommand\footnotetextcopyrightpermission[1]{}

\thanks{Accepted for publication at the ACM International Conference on Multimedia 2026.}
\begin{document}

\title{Scaling Vision-Language Models Is Not Enough to Mitigate Bias}
\author{Ioannis Sarridis}

\email{gsarridis@iti.gr}
\affiliation{%
  \institution{Information Technologies Institute, CERTH}
  \city{Thessaloniki}
  \country{Greece}
}

\author{Ioannis Kompatsiaris}
\email{ikom@iti.gr}
\affiliation{%
  \institution{Information Technologies Institute, CERTH}
  \city{Thessaloniki}
  \country{Greece}
}
\author{Symeon Papadopoulos}
\email{papadop@iti.gr}
\affiliation{%
  \institution{Information Technologies Institute, CERTH}
  \city{Thessaloniki}
  \country{Greece}
}
\renewcommand{\shortauthors}{Sarridis et al.}

\begin{abstract}

Vision-Language Models (VLMs) such as CLIP are now foundational to multimodal systems, yet their robustness to spurious correlations remains poorly understood at scale. We present the first large-scale empirical study of \nummodels{} publicly available VLMs, including 16 model families, covering a wide range of model sizes, 24 training datasets, and three evaluation benchmarks, namely ImageNet (overall performance), CelebA (typical single-attribute bias), and UrbanCars (complex multi-attribute biases).
Across these settings,  the Spearman correlation between model scale and performance weakens as evaluation shifts from ImageNet ($\rho{=}0.68$) to single-attribute ($\rho{=}0.48$) and further to multi-attribute ($\rho{=}0.05$) bias benchmarks.
In contrast, properties of the training data (size and quality) show more consistent relationships with worst-group accuracy across both bias benchmarks. Notably, curated datasets yield improvements of up to 25\% over uncurated alternatives at a comparable scale.
Finally, the effect of architectural choices (e.g., patch size, image resolution) is highly context-dependent, varying with the nature of the benchmark, including the type of bias and its spatial distribution within images.

\end{abstract}
 
\maketitle
 
\section{Introduction}\label{sec:intro}

\begin{figure}
    \centering
    \includegraphics[width=\linewidth]{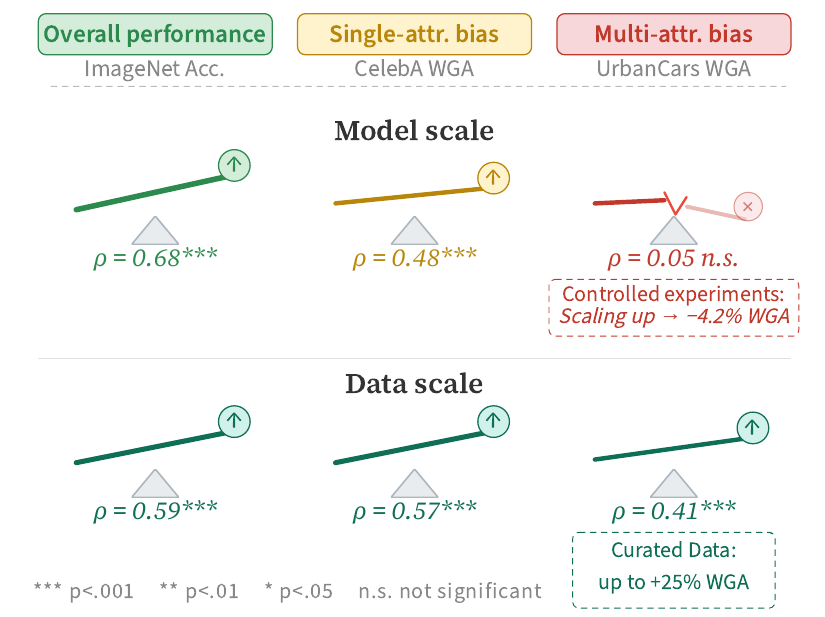}
    \caption{Model scale --- the dominant predictor of overall performance ($\rho{=}0.68$) --- becomes irrelevant for multi-attribute bias ($\rho{=}0.05$, n.s.), and controlled experiments show scaling \emph{reduces} Worst-Group Accuracy by $-4.2\%$ on average. Data scale retains predictive power across all bias settings ($\rho{=}0.59 \to 0.41$), with curated datasets improving WGA by up to $+25\%$ over unfiltered alternatives at matched scale.}
    \label{fig:teaser}
\end{figure}
Vision-Language Models (VLMs) such as CLIP~\cite{radford2021learning} and SigLIP~\cite{zhai2023sigmoid} have become foundational to modern multimodal systems, powering image retrieval, content moderation, recommendation engines, and text-to-image generation~\cite{jia2021scaling}. As these models are deployed at scale, their reliability becomes a practical concern, particularly their susceptibility to \emph{spurious correlations}, where models rely on shortcut features (e.g., image backgrounds, co-occurring objects, demographic attributes) rather than genuinely predictive ones~\cite{geirhos2020shortcut, sagawa2020distributionally,sarridis2024facex}. 
 
While spurious correlations have been extensively studied in supervised settings~\cite{sagawa2020distributionally, liu2021just, kirichenko2023last,sarridis2025flac, sarridis2025mavias}, the VLM setting introduces distinct challenges. VLMs are trained on web-scraped datasets containing billions of loosely curated image-text pairs, inheriting the biases of the Internet at unprecedented scale. Their training objectives align images and text in a shared embedding space, exploiting the co-occurrence patterns that serve as shortcuts. Moreover, zero-shot deployment means there is no task-specific correction stage; thus, whatever biases exist in the pre-trained representation are directly exposed to end users. Nowadays, there are hundreds of publicly available VLMs varying along multiple design axes, yet no study has systematically mapped how these choices affect robustness to spurious correlations.
In practice, model selection is overwhelmingly driven by overall performance benchmarks,
which implicitly assumes that better-performing models are also more robust to biases. 

In this paper, we conduct the first large-scale study of spurious correlation robustness in VLMs. Our study covers \nummodels{} publicly available models with parameter counts from 63M to 3.6B from 16 architectural families, trained on 24 distinct data sources ranging from 13M to 12.8B samples. 
We measure each model along a progression of increasing bias complexity: ImageNet~\cite{deng2009imagenet} serves as a baseline capturing the overall recognition ability, CelebA~\cite{liu2015deep, sagawa2020distributionally} tests robustness to a single spurious attribute (gender correlating with hair color), and UrbanCars~\cite{li2023whac} tests robustness to two simultaneous spurious attributes (background and co-occurring objects both correlating with car type). 

The analysis results show that model performance progressively departs from established VLM scaling laws \cite{cherti2023reproducible} --- which predict consistent gains with increased model and data scale --- as bias complexity increases. We term this breakdown \emph{bias complexity sensitivity}.
\Cref{fig:teaser} shows that model scale --- the factor most strongly associated with ImageNet performance ($\rho{=}0.68$) --- drops to $\rho{=}0.48$ for single-attribute bias and becomes negligible for multi-attribute bias ($\rho{=}0.05$, n.s.). Controlled comparisons further confirm that when only model size varies, scaling yields $+2.6$\% ImageNet accuracy on average but \emph{reduces} UrbanCars Worst-Group Accuracy (WGA) by $-4.2$\%. Note that WGA serves as the bias-aware performance metric. Model scaling, the dominant lever for improving standard VLM performance, loses its effectiveness in datasets and settings with more complex biases.
In contrast, other design factors exhibit more stable behavior. Training data properties --- dataset size and curation quality --- maintain more stable associations with WGA across both bias settings. Data size retains a correlation of $\rho{=}0.41$ even on multi-attribute bias, and curated datasets outperform size-matched unfiltered alternatives by up to 25\% in WGA. We also observe systematic effects related to token granularity; increasing patch size from 16 to 32 harms WGA performance on UrbanCars by $6.8$\% in controlled comparisons. 
 
In sum, this paper makes the following contributions:
\begin{enumerate}[leftmargin=*, nosep]
    \item \textbf{First large-scale study of spurious correlation robustness in VLMs}: We systematically evaluate \nummodels{} publicly available VLMs from 16 model families, diverse model scales, and 24 training datasets, providing the most comprehensive analysis of robustness to spurious correlations in VLMs to date.
    
    \item \textbf{Identification of \emph{bias complexity sensitivity}}: We show that model performance progressively departs from established VLM scaling laws as bias complexity increases. 

    \item \textbf{Insights into design factors for robust VLMs}: We show that training data properties (size and curation) are more reliable predictors of robustness, and quantify the impact of several VLM design factors on the model's robustness to spurious correlations.
\end{enumerate}
 The implementation of our experimental study is available online: \url{https://github.com/gsarridis/vlm-spurious-robustness}.
\section{Related Work}\label{sec:related}
 
\paragraph{Shortcut learning and spurious correlations}
Neural networks exploit spurious correlations --- features predictive in the training distribution but unreliable under distribution shift~\cite{geirhos2020shortcut}. This has been documented in image classification (e.g., models relying on backgrounds rather than objects~\cite{beery2018recognition, xiao2021noise}), natural language inference~\cite{mccoy2019right}, and visual question answering~\cite{agrawal2018don}. A rich body of work addresses mitigation~\cite{sagawa2020distributionally,arjovsky2019invariant,liu2021just, creager2021environment,kirichenko2023last,sarridis2025badd,sarridis2025vb}. 
In this work, rather than suggesting a bias mitigation methodology, we ask 
 which pre-training design choices inherently yield less biased models.
 
\paragraph{VLM scaling laws and evaluation.}
The CLIP model family~\cite{radford2021learning} established that contrastive language-image pre-training yields strong zero-shot classifiers. OpenCLIP~\cite{cherti2023reproducible} demonstrated reproducible scaling laws showing that both data volume and model size predict zero-shot \emph{average} accuracy.  ELEVATER~\cite{li2022elevater} provides multi-task evaluation of language-augmented visual models.

\paragraph{Bias and fairness in VLMs}
Demographic biases in CLIP embeddings have been documented through systematic probing~\cite{agarwal2021evaluating} and shown to propagate to downstream retrieval and generation systems~\cite{wang2021assessing}. 
Fang et al.~\cite{fang2022data} explored the impact of training data composition on CLIP's robustness to natural distribution shifts.
The ``accuracy on the line'' phenomenon~\cite{miller2021accuracy, taori2020measuring} --- where in-distribution and out-of-distribution accuracy are linearly correlated --- has been documented for distribution shifts w.r.t. a single attribute. 
We extend this line to the ecosystem scale, auditing \nummodels{} models and systematically relating design choices to worst-group performance.
 
\paragraph{Data-centric AI}
The importance of training data quality over quantity has gained recognition across machine learning~\cite{zha2023datacentric}. In the VLM setting, DFN~\cite{fang2023data} trains data filtering networks to select high-quality image-text pairs, and DFNDR~\cite{faghri2025mobileclip} extends this with dataset reinforcement.
DataComp~\cite{gadre2024datacomp} introduced controlled benchmarks for data curation, showing that filtering strategies outperform raw scaling for average accuracy.

\section{Methodology}\label{sec:setup}

\subsection{Problem Formulation}\label{sec:formulation}
 
A VLM consists of an image encoder $\phi_I$ and a text encoder $\phi_T$ that map inputs to a shared embedding space. Given a downstream classification task with label space $\mathcal{Y} = \{y_1, \ldots, y_C\}$, zero-shot inference is performed by constructing a text prompt $t_c$ for each class $c$ (e.g., ``a photo of a \{class\}''), encoding it as $\phi_T(t_c)$, and assigning an image $x$ to the class with maximum similarity:
\begin{equation}
    f(x) = \arg\max_{c \in \mathcal{Y}} \; \text{sim}\!\left(\phi_I(x), \; \phi_T(t_c)\right)
    \label{eq:zeroshot}
\end{equation}
where $\text{sim}(\cdot, \cdot)$ denotes cosine similarity. 
 
In this setting, suppose the data also carries $M$ spurious attributes $\mathcal{A} = \{A_1, \ldots, A_M\}$, where each $A_m$ takes values in a finite set (e.g., background type $\in \{\text{urban}, \text{country}\}$). A spurious attribute correlates with the target label in the training distribution but is not causally related to it. The combination of label and all spurious attributes induces a group structure $\mathcal{G} = \mathcal{Y} \times A_1 \times \cdots \times A_M$, partitioning the data into $K = |\mathcal{Y}| \cdot \prod_{m=1}^{M} |A_m|$ subgroups. When $M{=}1$ (e.g., CelebA: gender as spurious attribute), this yields $K{=}|\mathcal{Y}| \cdot |A_1|$ groups. When $M{=}2$ (e.g., UrbanCars: background \emph{and} co-occurring objects), the group count grows combinatorially to $K{=}|\mathcal{Y}| \cdot |A_1| \cdot |A_2|$. This combinatorial growth as $M$ increases is what makes multi-attribute bias fundamentally harder, since the zero-shot classifier $f$ must resist all $M$ shortcuts simultaneously.
 
Our goal is to understand how the design choices behind $\phi_I$ and $\phi_T$ (i.e., architecture, training data, model scale, and input representation) affect the zero-shot classifier's worst-group behavior under spurious correlations of varying complexity.
 
\subsection{Metrics}\label{sec:metrics}
 
We evaluate VLMs along two dimensions:
 
\paragraph{Overall performance.} Zero-shot ImageNet top-1 accuracy~\cite{deng2009imagenet} serves as the standard proxy for visual recognition quality and the metric on which VLM scaling laws are typically established~\cite{cherti2023reproducible}.
 
\paragraph{Robustness to spurious correlations.} For a benchmark with group structure $\mathcal{G}$ induced by $M$ spurious attributes, we measure the WGA of the zero-shot classifier $f$ (\Cref{eq:zeroshot}):
\begin{equation}
    \text{WGA}(f) = \min_{(y, a_1, \ldots, a_M) \in \mathcal{G}} \frac{1}{|\mathcal{D}_g|} \sum_{(x, y') \in \mathcal{D}_g} \mathbf{1}[f(x) = y']
    \label{eq:wg}
\end{equation}
where $\mathcal{D}_g$ is the subset of test data belonging to group $g$. WGA captures the accuracy on the subgroup most affected by spurious correlations~\cite{sagawa2020distributionally}. 
 

\subsection{Benchmarks}\label{sec:benchmarks}
 
We select two benchmarks that differ in \emph{spurious correlation complexity}:

 \paragraph{CelebA~\cite{liu2015deep,sagawa2020distributionally}} (\textbf{single-attribute bias}).
This benchmark evaluates hair-color classification (blonde vs.\ non-blonde) on celebrity face images, where gender serves as a single spurious attribute. The $K{=}4$ subgroups are defined by hair color $\times$ gender. The minority subgroup ``blonde male'' has the lowest representation, and the WGA is typically attained on this group. 
 
\paragraph{UrbanCars~\cite{li2023whac}} (\textbf{multi-attribute biases}).
This benchmark evaluates car-type classification (urban vs.\ country) on synthetically composed images where \emph{two} spurious features are simultaneously controlled: the background scene (urban or country) and a co-occurring object (e.g., a traffic light for urban, a cow for country). Images are created by pasting car objects and co-occurring objects onto backgrounds, yielding $K{=}8$ subgroups from the $2 \times 2 \times 2$ combination of (object class) $\times$ (background) $\times$ (co-occurring object). Achieving high WGA requires resisting \emph{both} shortcuts simultaneously, i.e., a model must correctly classify a country car placed on an urban background with urban co-occurring objects, and vice versa. UrbanCars is considered a much more challenging benchmark than CelebA, due to both its nature and the number of spurious correlations involved \cite{li2023whac}.

\subsection{Model Collection}\label{sec:model_collection}
We evaluate \nummodels{} publicly available VLMs, all sourced from the OpenCLIP repository\footnote{\url{https://github.com/mlfoundations/open_clip}}. 
The collection involves 16 model families: 
CLIP ViT ($n{=}85$) \cite{radford2021learning}, 
ResNet\footnote{For convolutional families, the family name refers to the vision encoder only.} ($n{=}16$) \cite{radford2021learning}, 
ViTamin ($n{=}15$) \cite{chen2024vitamin}, 
SigLIP2 ($n{=}15$) \cite{tschannen2025siglip2}, 
ConvNeXt ($n{=}12$) \cite{liu2022convnet}, 
SigLIP ($n{=}11$) \cite{zhai2023sigmoid}, 
CLIPA ($n{=}7$) \cite{li2023clipa}, 
MobileCLIP2 ($n{=}6$) \cite{faghri2025mobileclip}, 
PE ($n{=}5$) \cite{bolya2025perception}, 
EVA ($n{=}5$) \cite{fang2023eva}, 
MobileCLIP ($n{=}4$) \cite{vasu2024mobileclip}, 
NLLB-SigLIP ($n{=}4$) \cite{visheratin2023nllb}, 
CoCa ($n{=}4$) \cite{yu2022coca}, 
and three smaller families ($n{\leq}2$ each: 
XLM-RoBERTa-CLIP \cite{conneau2020xmlroberta}, 
NLLB-CLIP \cite{visheratin2023nllb}, 
RoBERTa-CLIP \cite{liu2019roberta}). These models consist of 63M to 3.6B trainable parameters, use image resolutions from 224 to 512\,px, and patch sizes of 14, 16, and 32\,px. Note that we include all models available in OpenCLIP except for EVA02-E-14 and EVA02-E-14-plus due to computational resource limitations. All experiments were conducted on a single NVIDIA RTX 3090 Ti GPU.


 \subsection{Datasets}
 The considered model collection involves 24 training datasets. 
 
 \textbf{YFCC-15M}~\cite{thomee2016yfcc100m} and \textbf{Conceptual-12M}~\cite{changpinyo2021conceptual} are earlier web-crawled datasets of comparatively modest scale.
 \textbf{OpenAI-400M}~\cite{radford2021learning} is the proprietary dataset used to train the original CLIP models, consisting of 400M image-text pairs curated from the web.
 The \textbf{LAION} family~\cite{schuhmann2022laion} comprises large-scale, loosely filtered web-scraped corpora: LAION-400M, LAION-2B, LAION-5B, and LAION-Aesthetics-900M (a quality-filtered subset of LAION-2B retaining images with high predicted aesthetic scores). \textbf{Merged-2B} is a composite dataset combining LAION-2B and COYO-700M~\cite{kakaobrain2022coyo700m}, used to train the EVA model family~\cite{fang2023eva}.  
\textbf{WebLI-10B}~\cite{chen2023pali} is Google's large-scale multilingual image-text dataset, consisting of 10B image-text pairs crawled from the web across 109 languages, used to train the SigLIP and SigLIP2 model families. 
 The \textbf{CommonPool} and \textbf{DataComp} families~\cite{gadre2024datacomp} are derived from the DataComp benchmark, which provides a controlled testbed for data curation research. CommonPool refers to unfiltered candidate pools at various scales (13M, 128M, 1B, and 12.8B samples), while DataComp refers to filtered subsets of the same pools at matched scales (13M, 128M, 1B, and 12.8B), selected using the CLIP-score filtering strategy proposed in~\cite{gadre2024datacomp}. 
 \textbf{CommonCrawl-2.5B} is the large web-crawl corpus and \textbf{MetaCLIP}~\cite{xu2023demystifying} variants use curated subsets of CommonCrawl obtained via metadata-driven balancing, designed to replicate the distributional properties of the original OpenAI training data. The \textbf{MetaCLIP2-2.5B}~\cite{tian2025metaclip2} extends this approach with an updated curation pipeline and broader coverage. 
 \textbf{DFN-2B} and \textbf{DFN-5B}~\cite{fang2023data} are curated datasets produced by a learned Data Filtering Network (DFN) that scores and retains high-quality image-text pairs from large web crawls; DFN-5B is among the highest-quality datasets in our collection by average model performance. \textbf{DFNDR-2B}~\cite{faghri2025mobileclip} extends the DFN approach with dataset reinforcement.

\subsection{Evaluation Protocol}\label{sec:protocol}
 
All models are evaluated in the zero-shot setting formalized in \Cref{eq:zeroshot}. 

\paragraph{ImageNet} We follow the standard zero-shot protocol from~\cite{radford2021learning}, using the 80 prompt templates provided by OpenCLIP (e.g., ``a photo of a \{class\}'', ``a bad photo of the \{class\}'', ``a sculpture of the \{class\}'') and reporting top-1 accuracy over the 1,000 ImageNet classes.
 
 

 \paragraph{UrbanCars} The task is binary classification into \emph{urban} vs.\ \emph{country} car types. Rather than using the group labels directly, we use subtype descriptors to avoid leakage of spurious attributes (background, co-occurring object) into the text prompts. Prompts follow the template ``a photo of a \{subtype\} car'', with subtypes: \{\emph{compact, sports, sedan}\} for urban and \{\emph{truck, jeep, pickup}\} for country.

\paragraph{CelebA} The task is binary classification into \emph{blonde} vs.\ \emph{non-blonde} hair color. Prompts follow the template ``a photo of a person with \{descriptor\} hair'', with descriptors: \{\emph{blonde, light blonde, golden, platinum blonde}\} for blonde and \{\emph{dark, black, brown, red, grey, auburn}\} for non-blonde. For the latter, we also consider ``a photo of a \{\textit{brunette}, \textit{bald}\} person'' prompts. The asymmetry in prompt count reflects the greater diversity of the non-blonde category.

All evaluations use the models' native image resolution. Per model we report three metrics: ImageNet top-1 accuracy (overall performance), CelebA WGA (single-attribute bias robustness), and UrbanCars WGA (multi-attribute bias robustness).

\section{Results}
\subsection{Overall Performance vs. Robustness to Spurious Correlations}\label{sec:gap}

The common practice of selecting VLMs by ImageNet accuracy implicitly assumes that higher overall performance translates to greater robustness. We test this assumption directly and find it holds only partially for single-attribute biases and breaks down for multi-attribute ones.

The Spearman correlation between ImageNet top-1 accuracy and WGA is $\rho{=}0.68$ ($p{<}10^{-6}$) for CelebA but only $\rho{=}0.27$ ($p{=}0.0002$) for UrbanCars. Improvements in overall accuracy translate into improved robustness to single-attribute biases, but this relationship largely dissolves for the more complex multi-attribute setting. For instance, ViT-gopt-16-SigLIP2-384 (85.0\% acc. on ImageNet) achieves 73.6\% UrbanCars WGA; ViT-SO400M-14-SigLIP-384 (83.1\% acc. on ImageNet) achieves 86.4\% UrbanCars WGA; and ViT-bigG-14-worldwide-378 (83.0\% acc. on ImageNet) achieves only 31.2\% UrbanCars WGA. Models within a 2\% ImageNet accuracy range thus span over 55\% in multi-attribute biases.



\begin{table}[t]
\centering
\caption{Spearman $\rho$ between design factors and performance across three benchmarks. 
$^{***}p{<}.001$; $^{**}p{<}.01$; $^{*}p{<}.05$; n.s.\ not significant.}
\label{tab:correlations}
\small
\resizebox{\linewidth}{!}{%

\begin{tabular}{@{}lccc@{}}
\toprule
& \textbf{ImageNet acc.} & \textbf{CelebA WGA} & \textbf{UrbanCars WGA} \\
& \textit{overall} & \textit{single-attr.} & \textit{multi-attr.} \\
\midrule
\multicolumn{4}{@{}l}{\emph{Scale}} \\
\quad Parameters  & $+0.68^{***}$ & $+0.48^{***}$ & $+0.05~\text{n.s.}$ \\
\quad Data size   & $+0.59^{***}$ & $+0.57^{***}$ & $+0.41^{***}$ \\
\midrule
\multicolumn{4}{@{}l}{\emph{Representation}} \\
\quad Tokens      & $+0.70^{***}$ & $+0.52^{***}$ & $+0.27^{***}$ \\
\quad Patch size  & $-0.60^{***}$ & $-0.38^{***}$ & $-0.22^{**}$ \\
\quad Image size  & $+0.42^{***}$ & $+0.32^{***}$ & $+0.19^{**}$ \\
\bottomrule
\end{tabular}}
\end{table}

Having established that overall accuracy is a poor proxy for robustness to multi-attribute biases, we next examine which design factors predict performance on each benchmark and how those relationships change with bias complexity.
Reading \Cref{tab:correlations} rows from left to right, the correlation between a given design factor and performance \emph{decays} as bias complexity increases. The decay is most dramatic for model scale, where the parameter count drops from $\rho{=}+0.68$ on ImageNet to $\rho{=}+0.05$ on UrbanCars WGA. Representation factors (number of visual tokens, patch size, and image resolution) also decay, though they retain significance on both bias benchmarks. Token count exhibits a stronger correlation than patch size or image resolution individually because it is a function of both ($\text{tokens} = (\text{image size} / \text{patch size})^2$), amplifying their joint effect.

Training data size is the factor that resists this decay most strongly, holding at $\rho{=}+0.59$ and $\rho{=}+0.57$ for ImageNet and CelebA WGA, before declining only modestly to $\rho{=}+0.41$ for UrbanCars WGA. Conventional scaling laws emphasize that both model size and data size must increase together for optimal performance~\cite{cherti2023reproducible}. The key departure from established VLM scaling laws is that when the target shifts from overall accuracy to robustness to bias, model size loses its predictive power while data size retains it.

However, these raw correlations reflect the joint influence of all design factors, and many covary. The following section disentangles these effects through controlled comparisons.

\subsection{Design Factors}\label{sec:gradient}

\begin{figure*}[t]
    \centering
    \begin{subfigure}[t]{0.33\textwidth}
        \centering
        \includegraphics[width=\textwidth,trim={0.0cm 0.0cm 0.0cm 0.0cm},clip]{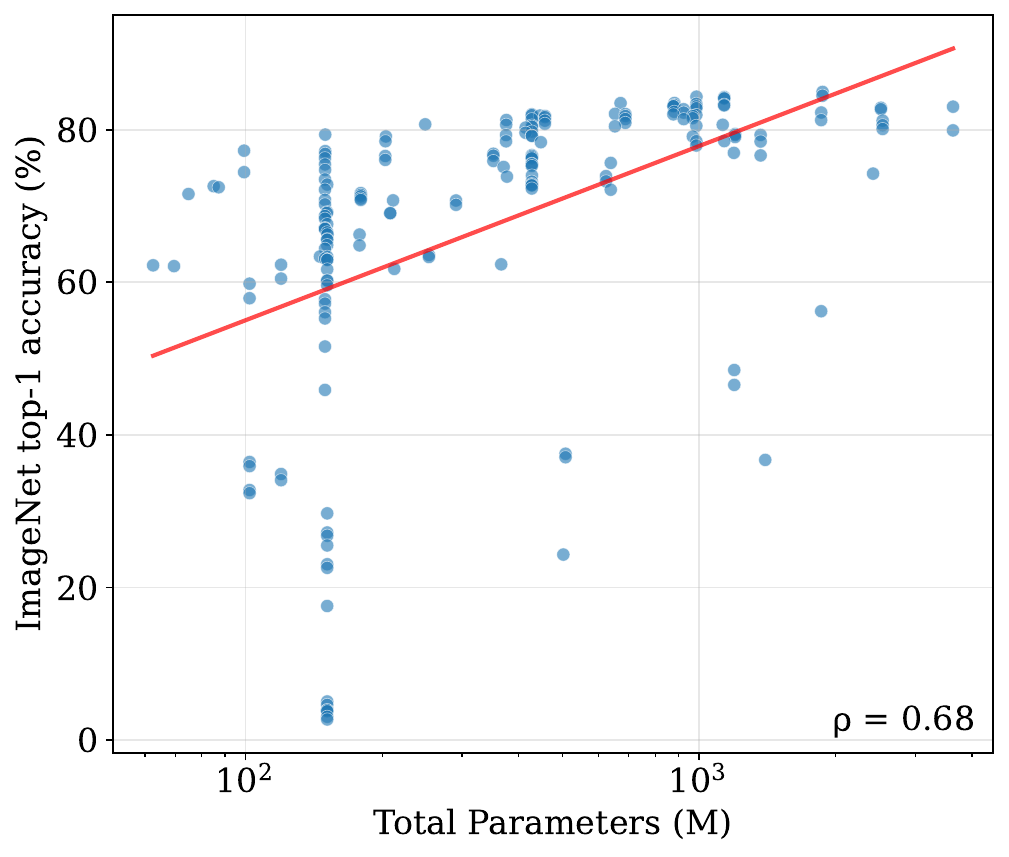}
        \caption{Params vs.\ ImageNet}
        \label{fig:scale_in}
    \end{subfigure}
    \hfill
    \begin{subfigure}[t]{0.33\textwidth}
        \centering
        \includegraphics[width=\textwidth,trim={0.0cm 0.0cm 0.0cm 0.0cm},clip]{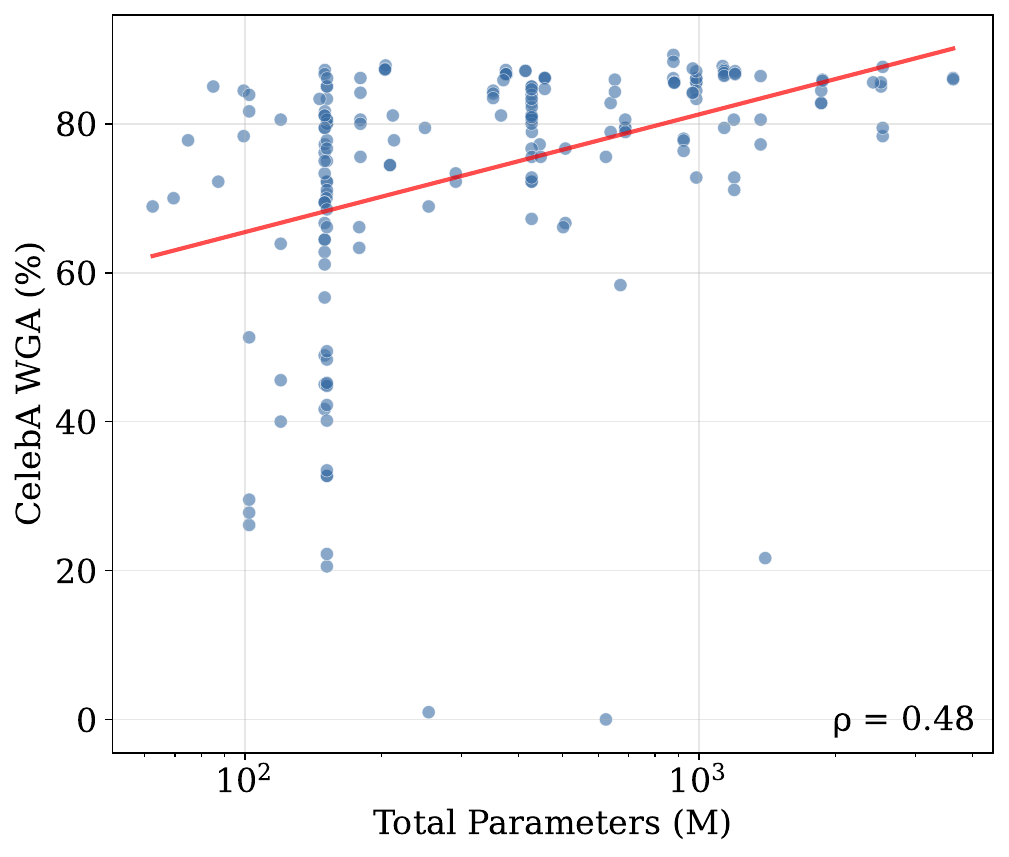}
        \caption{Params vs.\ CelebA WGA}
        \label{fig:scale_ca}
    \end{subfigure}
    \hfill
    \begin{subfigure}[t]{0.33\textwidth}
        \centering
        \includegraphics[width=\textwidth]{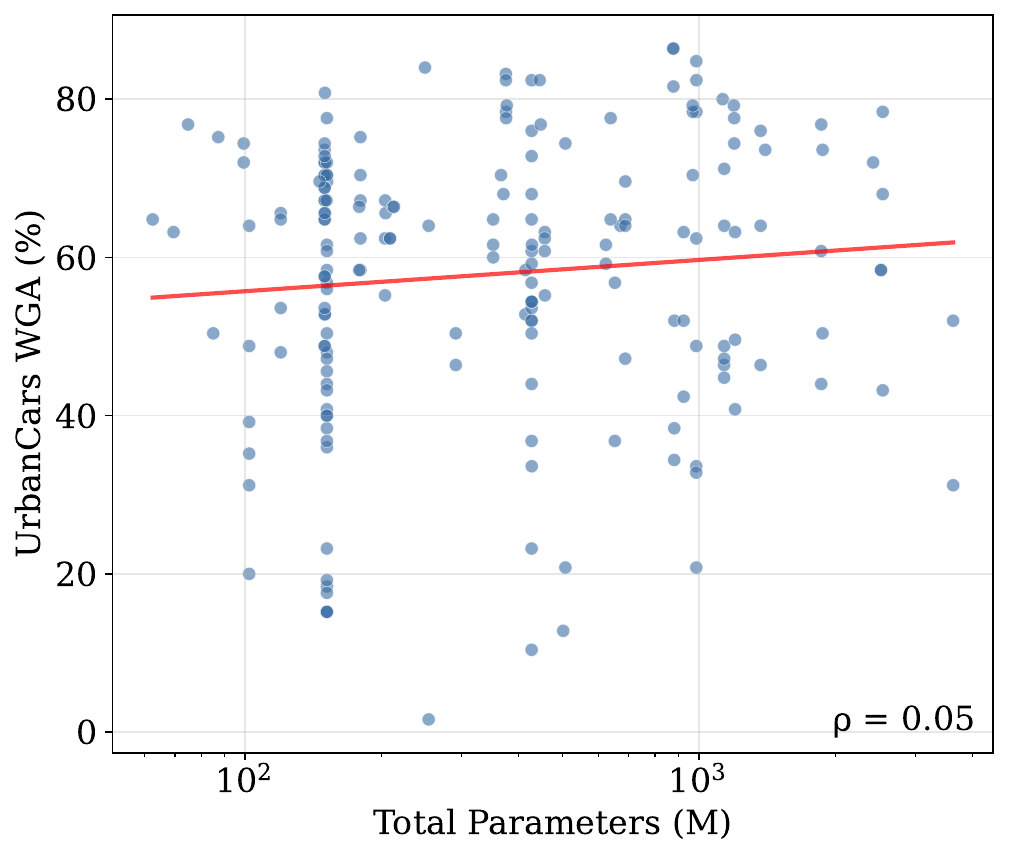}
        \caption{Params vs.\ UrbanCars WGA}
        \label{fig:scale_uc}
    \end{subfigure}
    \caption{\textbf{The bias complexity sensitivity through model scale.} The trend progressively flattens from left to right: parameters strongly predict overall performance but not robustness to multi-attribute biases. Points colored by model family.}
    \label{fig:scaling}
\end{figure*}

\noindent\textbf{Model Scale.}
This factor is the quintessential axis of VLM progress, and its relationship with performance across our three benchmarks illustrates the bias complexity sensitivity. \Cref{fig:scaling} displays three scatter plots side by side, where one may observe that the clear positive trend on ImageNet (left) progressively flattens through CelebA (center) to a near-uniform cloud on UrbanCars (right).
 
To move beyond correlations, we identify 11 matched groups where only the ViT backbone size varies while model family, training data, image resolution, and patch size are held constant. 
\begin{table*}[t]
\centering
\caption{Isolated model scale effect. Each row is a matched group where only ViT size varies. Arrows indicate accuracy at each size step. Scaling reliably improves ImageNet (25/26 positive) but is unreliable for robustness to bias.}
\label{tab:scale_isolation}
\small
\begin{tabular}{@{}lllll@{}}
\toprule
\textbf{Base Configuration} & \textbf{Sizes} & \textbf{ImageNet acc. (\%)} & \textbf{CelebA WGA (\%)} & \textbf{UrbanCars WGA (\%)} \\
\midrule
CLIP / LAION-2b / 224 / 14.0       & L$\to$H$\to$g$\to$bigG & 75.2$\to$77.9$\to$77.5$\to$80.1 & 83.3$\to$84.4$\to$78.9$\to$87.6 & 72.8$\to$48.8$\to$70.0$\to$78.4 \\
CLIP / CommonCrawl-2.5b / 224 / 14.0 & L$\to$H$\to$bigG & 77.9$\to$79.5$\to$80.9 & 84.4$\to$79.2$\to$78.9 & 48.0$\to$26.8$\to$55.6 \\
CLIPA / DataComp-1b / 224 / 14.0   & L$\to$H$\to$bigG & 79.6$\to$81.5$\to$82.7 & 87.1$\to$84.1$\to$85.6 & 52.8$\to$79.2$\to$58.4 \\
CLIPA / DataComp-1b / 336 / 14.0   & L$\to$H$\to$bigG & 80.3$\to$81.8$\to$82.9 & 87.1$\to$84.2$\to$85.0 & 58.4$\to$78.4$\to$58.4 \\
SigLIP2 / WebLI-10b / 256 / 16.0   & B$\to$L$\to$gopt & 79.3$\to$82.3$\to$84.5 & 86.7$\to$85.5$\to$85.9 & 82.4$\to$34.4$\to$50.4 \\
SigLIP2 / WebLI-10b / 384 / 16.0   & B$\to$L$\to$gopt & 80.7$\to$83.2$\to$85.0 & 86.7$\to$85.5$\to$85.8 & 78.4$\to$38.4$\to$73.6 \\
CLIP / MetaCLIP2-2.5b / 224 / 14.0 & H$\to$bigG & 73.1$\to$79.9 & 83.0$\to$86.1 & 46.1$\to$52.0 \\
CLIP / MetaCLIP2-2.5b / 378 / 14.0 & H$\to$bigG & 82.3$\to$83.0 & 84.4$\to$86.0 & 76.8$\to$31.2 \\
SigLIP / WebLI-10b / 256 / 16.0    & B$\to$L & 75.8$\to$80.4 & 86.6$\to$85.9 & 61.6$\to$36.8 \\
SigLIP / WebLI-10b / 384 / 16.0    & B$\to$L & 78.5$\to$82.1 & 87.4$\to$84.3 & 67.2$\to$56.8 \\
SigLIP2 / WebLI-10b / 512 / 16.0   & B$\to$L & 81.3$\to$83.5 & 86.7$\to$85.6 & 77.6$\to$52.0 \\
\midrule
\multicolumn{2}{@{}l}{\textbf{Mean effect of scaling up ($n{=}26$)}} & \textbf{+2.63\%} & \textbf{$-$0.56\%} & \textbf{$-$4.22\%} \\
\multicolumn{2}{@{}l}{Direction (positive / negative)} & 25 / 1 & 10 / 16 & 12 / 13 \\
\bottomrule
\end{tabular}
\end{table*}
 As reported in \Cref{tab:scale_isolation}, scaling yields $+2.63$\% on average on ImageNet, with 25 of 26 comparisons positive --- a near-universal benefit. On CelebA WGA, the mean effect reverses to $-0.56$\%, with 16 of 26 comparisons negative: scaling more often \emph{hurts} than helps robustness to single-attribute biases. On UrbanCars WGA, scaling produces $-4.22$\% on average, with outcomes split nearly evenly (12 positive, 13 negative) and large variance across configurations.

\begin{figure*}[t]
    \centering
    \begin{subfigure}[t]{0.33\textwidth}
        \centering
        \includegraphics[width=\textwidth,trim={0.0cm 0.0cm 0.0cm 0.0cm},clip]{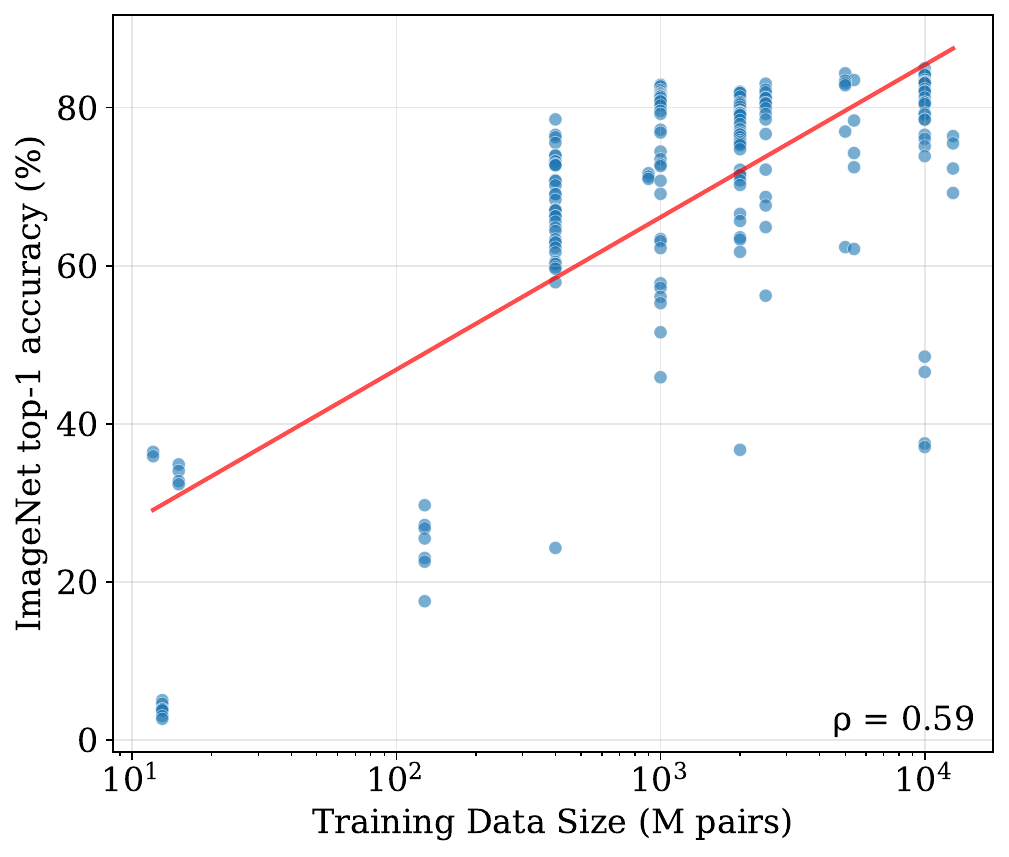}
        \caption{Data size vs.\ ImageNet}
        \label{fig:datasize_in}
    \end{subfigure}
    \hfill
    \begin{subfigure}[t]{0.33\textwidth}
        \centering
        \includegraphics[width=\textwidth,trim={0.0cm 0.0cm 0.0cm 0.0cm},clip]{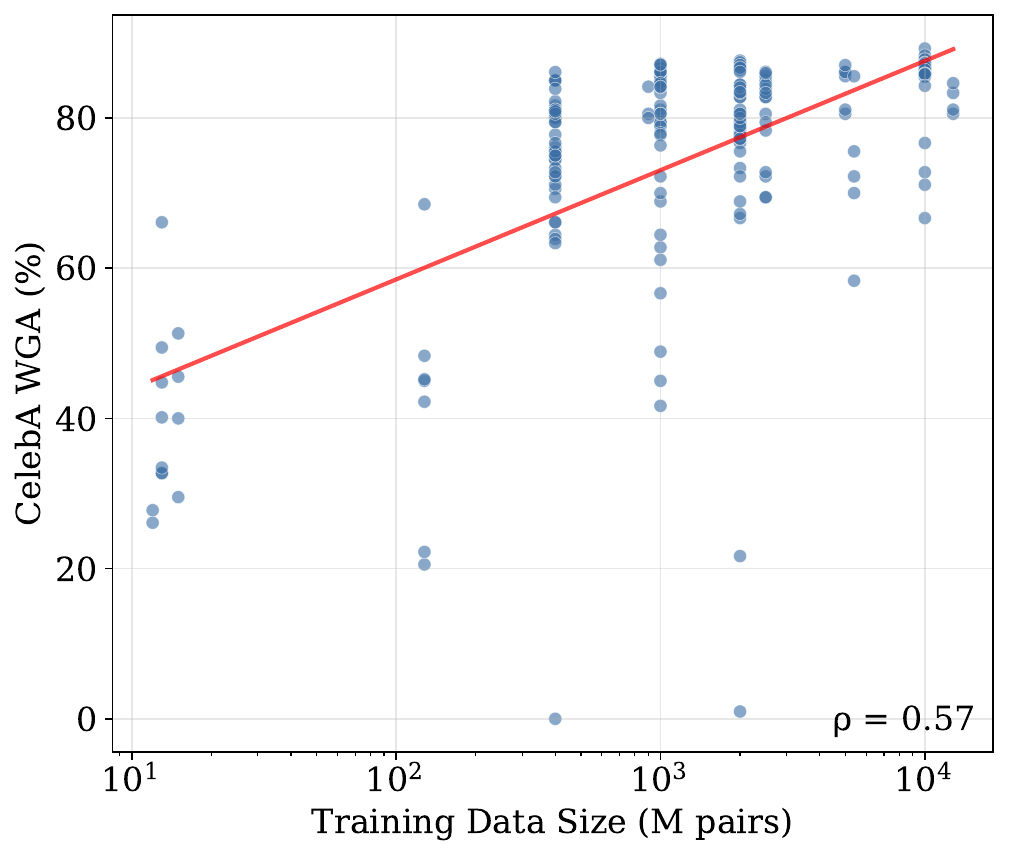}
        \caption{Data size vs.\ CelebA WGA}
        \label{fig:datasize_ca}
    \end{subfigure}
    \hfill
    \begin{subfigure}[t]{0.33\textwidth}
        \centering
        \includegraphics[width=\textwidth]{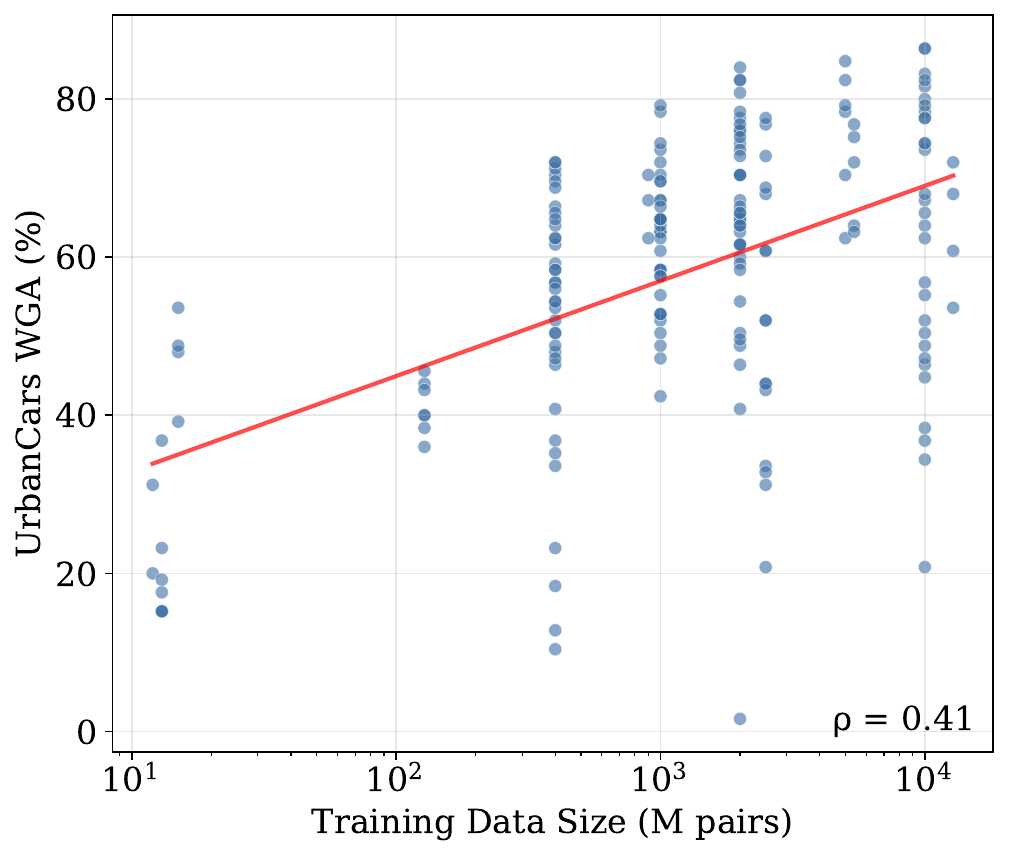}
        \caption{Data size vs.\ UrbanCars WGA}
        \label{fig:datasize_uc}
    \end{subfigure}
    \caption{\textbf{Training data size across benchmarks.} Unlike model parameters (\Cref{fig:scaling}), the positive trend persists on UrbanCars WGA, reflecting data size as the least sensitive design factor to bias complexity.}
    \label{fig:datasize_three}
\end{figure*}

\noindent\textbf{Training Data.}
The data used for pre-training is the strongest and most persistent predictor of WGA, both in terms of dataset size and curation quality.
As in \Cref{fig:scaling} for model-scaling, \Cref{fig:datasize_three} shows the trend for the training data size variable. 
Unlike model parameters, the positive trend does \emph{not} vanish on UrbanCars --- a clear slope persists across all three panels, confirming data size as the factor most resistant to the bias complexity sensitivity.

Beyond size, the \emph{quality} of training data matters substantially. \Cref{tab:data} presents dataset-level statistics, revealing that dataset rankings are highly benchmark-dependent. DFNDR-2B leads on UrbanCars WGA (78.7\%) but is mid-tier on ImageNet (78.4\%) and CelebA (78.0\%). DFN-5B achieves the highest CelebA WGA (86.2\%) and ImageNet accuracy (83.4\%). 
It is worth noting that among datasets of comparable scale ($\sim$2B samples), the curated DFNDR-2B achieves 78.7\% UrbanCars WGA on average, while the unfiltered CommonCrawl-2.5B achieves only 54.1\% --- a 25-point gap attributable to curation alone. 
Note that if we consider only the filtered/curated datasets, model scale remains a non-significant predictor of UrbanCars WGA with $\rho=0.10$.
\begin{table}[t]
\centering
\caption{Performance w.r.t. training datasets. Reported datasets involved in $\geq 3$ models. Sorted by UrbanCars WGA. IN, CA, and UC stand for ImageNet, CelebA, and UrbanCars, respectively.}
\label{tab:data}
\small
\resizebox{\linewidth}{!}{%

\begin{tabular}{@{}lrrr@{}}
\toprule
\textbf{Dataset} & \textbf{IN acc. (\%)} & \textbf{CA WGA (\%)} & \textbf{UC WGA (\%)} \\
\midrule
DFNDR-2B         & 78.4$\pm$3.7  & 78.0$\pm$2.4  & \textbf{78.7}$\pm$5.0  \\
DFN-5B           & \textbf{83.4}$\pm$0.7  & \textbf{86.2}$\pm$0.6  & 77.0$\pm$10.1 \\
MetaCLIP-5.4B    & 74.1$\pm$8.0  & 72.3$\pm$9.8  & 70.2$\pm$6.3  \\
LAION-A-900M     & 71.3$\pm$0.4  & 81.6$\pm$2.3  & 66.7$\pm$4.0  \\
WebLI-10B        & 75.9$\pm$13.8 & 84.7$\pm$5.4  & 63.2$\pm$17.8 \\
DFN-2B           & 79.9$\pm$3.0  & 73.3$\pm$7.8  & 62.8$\pm$10.8 \\
LAION-2B         & 71.8$\pm$9.6  & 76.0$\pm$21.1 & 62.2$\pm$16.2 \\
DataComp-1B      & 77.3$\pm$6.3  & 81.1$\pm$6.0  & 62.0$\pm$9.2  \\
CommonPool-1B    & 54.0$\pm$4.5  & 52.7$\pm$8.8  & 62.0$\pm$7.7  \\
CommonPool-12.8B & 74.7$\pm$2.2  & 82.1$\pm$2.2  & 60.8$\pm$7.2  \\
LAION-400M       & 67.4$\pm$5.5  & 79.0$\pm$4.8  & 60.3$\pm$8.8  \\
Merged-2B        & 78.6$\pm$2.6  & 78.7$\pm$5.9  & 57.0$\pm$8.5  \\
CommonCrawl-2.5B & 75.0$\pm$6.1  & 77.7$\pm$6.3  & 54.1$\pm$18.7 \\
MetaCLIP2-2.5B   & 77.4$\pm$10.5 & 84.2$\pm$1.5  & 49.7$\pm$17.3 \\
OpenAI-400M      & 65.2$\pm$11.5 & 70.6$\pm$18.2 & 48.7$\pm$15.6 \\
MetaCLIP-400M    & 69.1$\pm$5.3  & 74.7$\pm$7.5  & 47.5$\pm$25.4 \\
YFCC-15M         & 33.5$\pm$1.2  & 41.6$\pm$9.3  & 47.4$\pm$6.0  \\
CommonPool-128M  & 23.8$\pm$3.6  & 40.6$\pm$17.7 & 40.7$\pm$3.6  \\
CommonPool-13M   &  3.8$\pm$0.9  & 44.4$\pm$12.4 & 21.2$\pm$8.2  \\

\bottomrule
\end{tabular}}
\end{table}

\begin{figure*}[t]
    \centering
    \includegraphics[width=\textwidth]{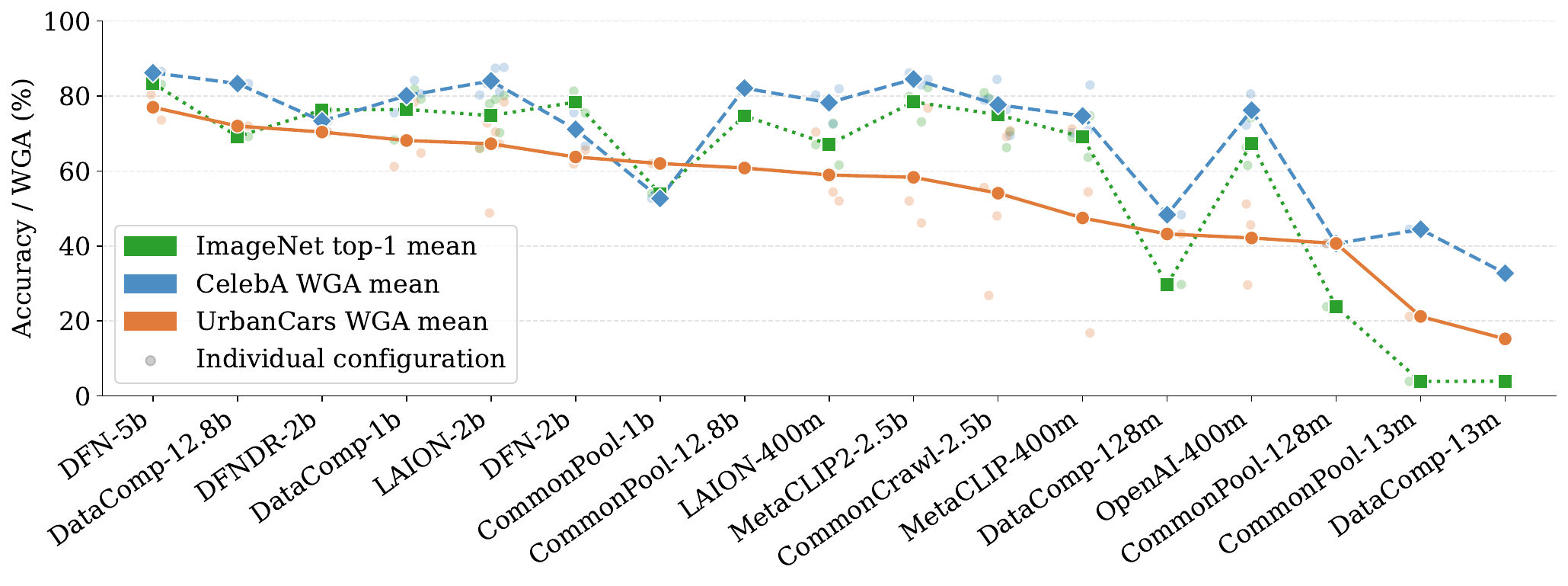}
    \caption{\textbf{Isolated training data effect: per-dataset accuracy across matched architecture groups.}
    Small semi-transparent dots show accuracy for each architecture configuration in which that dataset appears; lines connect the per-dataset means across all three benchmarks (ImageNet, CelebA, and UrbanCars).
    \textbf{Vertical spread} of dots at a given dataset reflects how much the effect depends on the architecture configuration.
    Datasets sorted by UrbanCars WGA mean.}
    \label{fig:data_isolation}
    \Description[<short description>]{<long description>}
\end{figure*}

To disentangle dataset effects from confounding architectural choices, we identify 7 matched groups where only the training dataset varies while model family, image resolution, and patch size remain constant. \Cref{fig:data_isolation} visualizes these results as a per-dataset dot plot, where each dataset appears on the horizontal axis and individual dots represent accuracy under different architecture configurations. Consistent with the earlier findings, we observe that curated and larger datasets tend to exhibit higher robustness.

\begin{figure*}[t]
    \centering

    \begin{subfigure}[t]{0.39\textwidth}
        \centering
        \includegraphics[width=\linewidth,trim={0.0cm 0.0cm 0.0cm 0.0cm}]{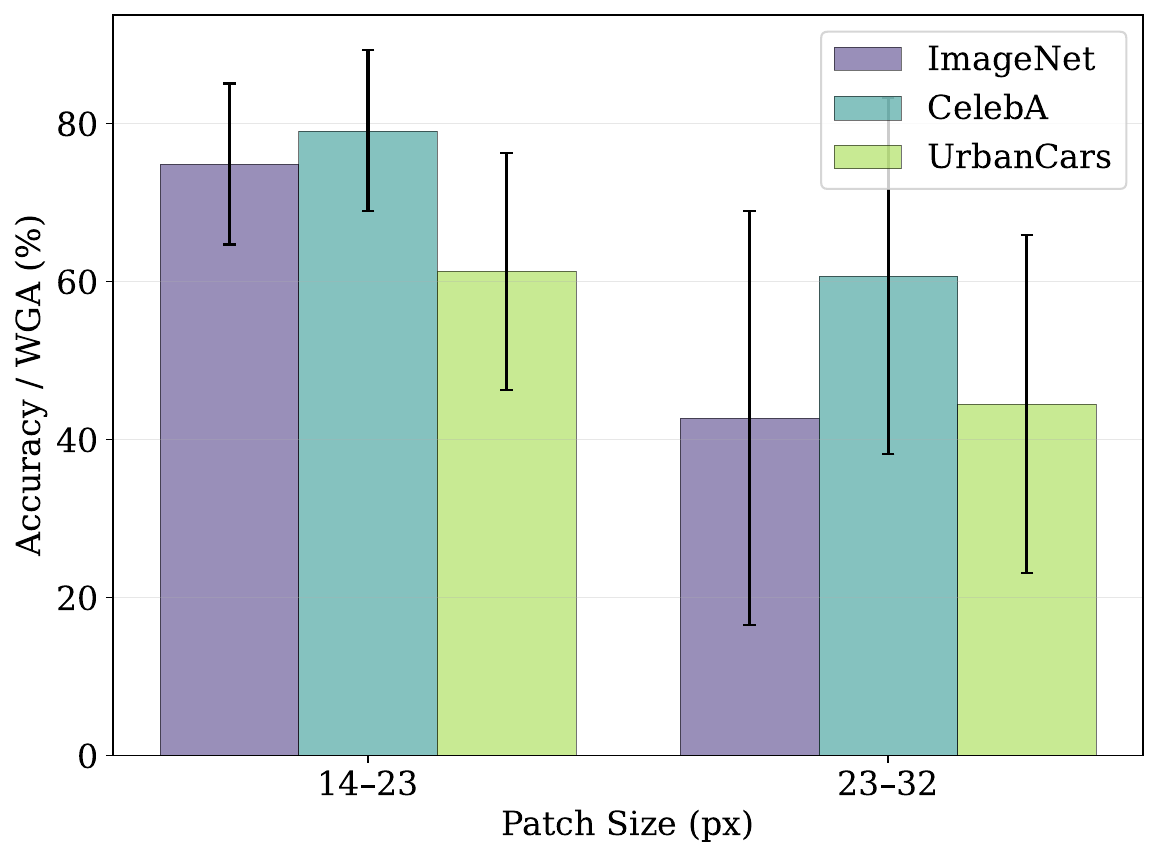}
        \caption{}
        \label{fig:patchsize_bars}
    \end{subfigure}
    \hfill
    \begin{subfigure}[t]{0.60\textwidth}
        \centering
        \includegraphics[width=\linewidth,trim={0.0cm 0.1cm 0.0cm 0.0cm}]{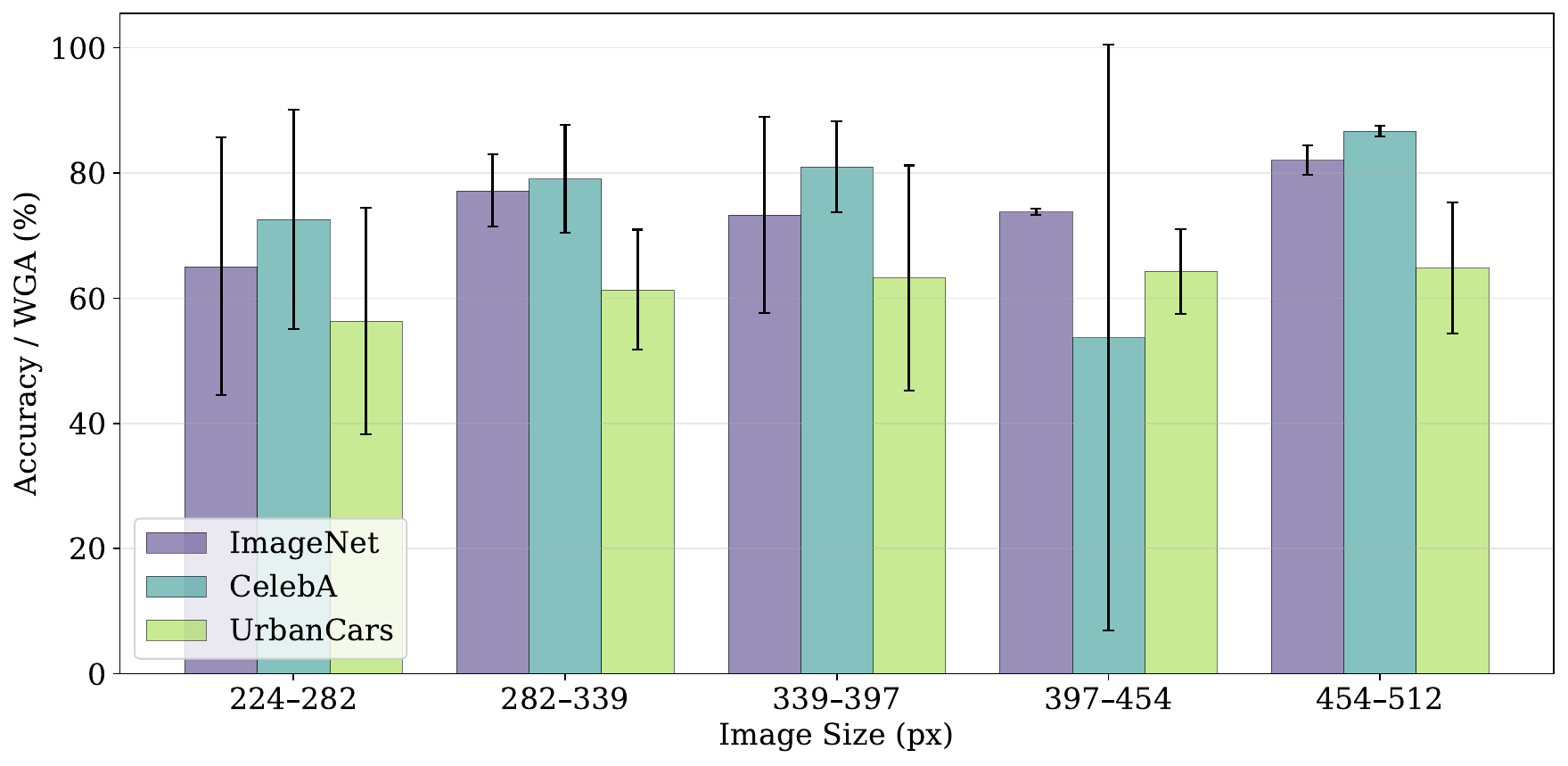}
        \caption{}
        \label{fig:imagesize_bars}
    \end{subfigure}

    \caption{\textbf{Effect of patch and image size on model performance. Higher resolution is associated with improved performance across all benchmarks, though the effect is modest compared to patch size.}}
    \label{fig:patch_and_resolution}
\end{figure*}

\noindent\textbf{Patch Size and Image Resolution.}
Tokenization granularity --- the number of visual tokens processed by the model, determined jointly by patch size and image resolution --- is the second factor that retains predictive power across bias benchmarks. \Cref{fig:patchsize_bars} and \Cref{fig:imagesize_bars} present the effects of patch size and image resolution, respectively. Smaller patches (patch-14 vs.\ patch-32) are associated with higher accuracy across all three benchmarks, with the gap most pronounced on ImageNet ($+$34.4\%). Similarly, higher image resolution shows a positive trend as well.


However, comparisons confound patch size and image resolution with other design choices (e.g., models with smaller patches tend to be newer, larger, and trained on better data). We therefore turn to controlled comparisons.

In 7 matched groups where only patch size varies (holding architecture, training data, and ViT backbone size constant), the results reveal a benchmark-dependent effect. As shown in \Cref{tab:patchsize_isolation}, increasing patch size from 16 to 32 reduces UrbanCars WGA by $6.8$\% on average, with 6 of 7 comparisons negative. 

\begin{table}[t]
\centering
\caption{Isolated patch size effect for 7 matched groups. Increasing the patch size from 16 to 32 predominantly hurts UrbanCars WGA ($-6.8$\% on average) but slightly helps CelebA WGA ($+1.6$\%), revealing a benchmark-dependent effect that is masked in aggregate statistics.}
\label{tab:patchsize_isolation}
\small
\resizebox{\linewidth}{!}{%
\begin{tabular}{@{}l ccc ccc ccc@{}}
\toprule
& \multicolumn{3}{c}{\textbf{IN acc. (\%)}} & \multicolumn{3}{c}{\textbf{CA WGA (\%)}} & \multicolumn{3}{c}{\textbf{UC WGA (\%)}} \\
\cmidrule(lr){2-4} \cmidrule(lr){5-7} \cmidrule(lr){8-10}
\textbf{Base Configuration} & p16 & p32 & $\Delta$ & p16 & p32 & $\Delta$ & p16 & p32 & $\Delta$ \\
\midrule
CLIP/LAION-400M/B        & 68.0 & 61.6 & $-$6.4 & 77.4 & 81.9 & $+$4.5 & 66.4 & 52.0 & $-$14.4 \\
SigLIP2/WebLI-10B/B      & 79.9 & 73.8 & $-$6.1 & 86.8 & 86.7 & $-$0.1 & 80.4 & 79.2 & $-$1.2  \\
CLIP/CommonCrawl-2.5B/B  & 70.4 & 66.2 & $-$4.2 & 69.4 & 76.4 & $+$7.0 & 70.8 & 69.2 & $-$1.6  \\
CLIP/MetaCLIP-400M/B     & 69.0 & 63.6 & $-$5.4 & 70.3 & 70.8 & $+$0.5 & 71.2 & 54.4 & $-$16.8 \\
CLIP/OpenAI-400M/B       & 66.4 & 61.5 & $-$4.9 & 72.2 & 75.8 & $+$3.6 & 51.2 & 29.6 & $-$21.6 \\
CLIP/DataComp-1B/B       & 68.3 & 72.8 & $+$4.5 & 75.6 & 72.2 & $-$3.4 & 61.2 & 70.4 & $+$9.2  \\
CLIP/LAION-2B/B          & 70.2 & 66.1 & $-$4.1 & 81.1 & 80.3 & $-$0.8 & 67.2 & 66.0 & $-$1.2  \\
\midrule
\textbf{Mean $\Delta$}   & \multicolumn{2}{c}{} & $-$3.8 & \multicolumn{2}{c}{} & $+$1.6 & \multicolumn{2}{c}{} & $-$6.8 \\
\bottomrule
\end{tabular}}
\end{table}
In contrast to UrbanCars, the controlled comparisons on CelebA do not reveal a clear effect of patch size, with both positive and negative changes observed across configurations and only a small average improvement (+1.6\%). This suggests that the influence of tokenization granularity is likely benchmark-dependent and may interact with multiple factors such as the target task, the type of spurious correlations, and the visual characteristics of the images (e.g., close-up facial images versus scene-centric images).

Regarding the image resolution, as reported in \Cref{tab:imagesize_isolation}, across 25 matched controlled comparisons, we observe modestly positive effects overall. Increasing resolution yields $+3.39$\% on UrbanCars WGA on average, consistent with the hypothesis that additional spatial detail helps resist scene-level shortcuts. On CelebA, the isolated effect is close to zero ($-0.33$\%), and on ImageNet it is small but positive ($+1.73$\%).

\begin{table}[t]
\centering
\caption{Isolated image resolution effect across all three benchmarks. Mean accuracy (or WGA) change when increasing resolution from the baseline. ImageNet gains are the most consistent; UrbanCars benefits inconsistently; CelebA is unaffected.}
\label{tab:imagesize_isolation}
\small
\resizebox{\linewidth}{!}{%
\begin{tabular}{@{}lrrr@{}}
\toprule
\textbf{Image Size} & \textbf{$\Delta$ IN acc. (\%)} & \textbf{$\Delta$ CA WGA (\%)} & \textbf{$\Delta$ UC WGA (\%)} \\
\midrule
224$\to$240 & $+$2.03 & $-$5.83 & $-$8.00 \\
224$\to$256 & $+$0.31 & $-$0.60 & $-$0.80 \\
224$\to$336 & $+$0.56 & $-$0.54 & $+$0.80 \\
224$\to$378 & $+$4.27 & $+$0.19 & $+$5.56 \\
224$\to$384 & $+$2.31 & $-$0.23 & 0.00 \\
224$\to$512 & $+$2.94 & 0.00 & $-$1.20 \\
256$\to$384 & $+$1.40 & $-$0.21 & $+$9.76 \\
256$\to$512 & $+$2.13 & $+$0.44 & $+$5.60 \\
384$\to$512 & $+$0.51 & $+$0.19 & $+$3.73 \\
\midrule
\textbf{Overall} & \textbf{+1.73\%} & \textbf{$-$0.33\%} & \textbf{+3.39\%} \\
\bottomrule
\end{tabular}}
\end{table}

\noindent\textbf{Model Families.}
\Cref{tab:arch} presents the full architecture-level ranking. The ranking is substantially benchmark-dependent: MobileCLIP2 leads on UrbanCars WGA (80.1\%) but drops to seventh on CelebA (78.8\%), while SigLIP dominates CelebA (87.0\%) but is mid-tier on UrbanCars (67.9\%). The families that perform consistently well across both bias benchmarks are SigLIP and CLIPA, which rank in the top half on both. 

\begin{table}[t]
\centering
\caption{Performance w.r.t. model families. Reported models with $\geq 3$ instances. Results sorted by UrbanCars WGA.}
\label{tab:arch}

\begin{tabular}{lccc}
\toprule
\textbf{Model}& \textbf{IN acc. (\%)} & \textbf{CA WGA (\%)} & \textbf{UC WGA (\%)} \\
\midrule
MobileCLIP2 & 78.8$\pm$3.9  & 78.8$\pm$1.2  & 80.1$\pm$3.7  \\
PE                    & 74.1$\pm$8.0  & 72.3$\pm$9.8  & 70.2$\pm$6.3  \\
SigLIP                & 79.7$\pm$2.9  & 87.0$\pm$1.4  & 67.9$\pm$15.2 \\
CLIPA                 & 81.1$\pm$1.5  & 85.8$\pm$1.4  & 65.1$\pm$10.7 \\
ConvNeXt              & 74.1$\pm$4.3  & 83.7$\pm$3.5  & 61.9$\pm$9.3  \\
NLLB-SigLIP           & 42.4$\pm$6.0  & 71.8$\pm$4.1  & 61.8$\pm$27.4 \\
ViTamin               & 77.6$\pm$7.3  & 79.8$\pm$5.3  & 60.9$\pm$8.1  \\
MobileCLIP            & 75.3$\pm$2.2  & 84.6$\pm$2.3  & 60.6$\pm$10.6 \\
SigLIP2               & 82.1$\pm$3.0  & 86.4$\pm$0.6  & 60.1$\pm$17.4 \\
EVA                   & 78.5$\pm$2.2  & 78.8$\pm$5.1  & 59.8$\pm$9.7  \\
CLIP                  & 62.0$\pm$23.2 & 71.0$\pm$16.5 & 53.3$\pm$18.3 \\
CoCa                  & 68.7$\pm$6.2  & 57.9$\pm$38.4 & 52.0$\pm$34.2 \\
ResNet                & 54.1$\pm$16.4 & 55.1$\pm$24.6 & 50.8$\pm$13.7 \\
\bottomrule
\end{tabular}
\end{table}

\subsection{Subgroup Analysis}

\begin{table}[h]
    \centering
    \small
        \caption{Spearman $\rho$ between scaling and performance on the UrbanCars subgroups.}
\begin{center}
\small
    \begin{tabular}{lcc}
\toprule
\textbf{Subgroups} &\textbf{Model Size} &\textbf{ Data Size} \\
\midrule
Fully bias-aligned & 0.52  & 0.53\\
Bias-conflicting w.r.t. background & 0.49  & 0.61\\
Bias-conflicting w.r.t. co-occurring object  & 0.25  & 0.50\\
Fully bias-conflicting  & 0.16  & 0.54\\
\bottomrule
\end{tabular}
\end{center}
    \label{tab:uc_subgroups}
\end{table}\
To better understand why model scale loses effectiveness on robustness benchmarks, we analyze the UrbanCars subgroups by the degree of conflict between the target label and spurious attributes. We define bias-aligned samples as those where both spurious attributes (background and co-occurring object) agree with the target class, and bias-conflicting samples as those where one or both contradict it. As shown in \Cref{tab:uc_subgroups}, the correlation between model size and WGA decreases monotonically as the number of conflicting attributes increases, dropping from $\rho=0.52$ for fully bias-aligned samples to $\rho=0.16$ for fully bias-conflicting ones. In contrast, training data size remains consistently correlated with performance across all subgroups.
Thus, scaling appears to improve context exploitation, which enhances performance on bias-aligned samples but becomes unreliable when multiple spurious attributes conflict with the target.


\subsection{Pareto-Optimal Models}\label{sec:anatomy}

The analyses in \Cref{sec:gradient} identify which factors matter on average. 
We now examine specific model subsets to illustrate how these factors interact in practice. 
We identify the Pareto-optimal models on the joint objective of maximizing ImageNet accuracy, UrbanCars WGA, and CelebA WGA simultaneously.

\begin{table}[t]
\centering
\caption{The 7 models on the 3-way Pareto frontier (ImageNet acc. $\times$ UrbanCars WGA $\times$ CelebA WGA). }
\label{tab:pareto3way}
\small
\resizebox{\linewidth}{!}{%
\begin{tabular}{lccc}
\toprule
\textbf{Model} & \textbf{IN (\%)} & \textbf{CA (\%)} & \textbf{UC (\%)} \\
\midrule
ViT-SO400M-14-SigLIP-384\_webli   & 83.1 & 89.2 & 86.4 \\
ViT-H-14-quickgelu\_dfn5b          & 83.4 & 87.0 & 84.8 \\
ViT-H-14-378-quickgelu\_dfn5b      & 84.3 & 86.1 & 78.4 \\
ViT-SO400M-16-SigLIP2-512\_webli   & 84.3 & 86.6 & 64.0 \\
ViT-SO400M-16-SigLIP2-384\_webli   & 84.2 & 87.3 & 48.8 \\
ViT-gopt-16-SigLIP2-256\_webli     & 84.5 & 85.9 & 50.4 \\
ViT-gopt-16-SigLIP2-384\_webli     & 85.0 & 85.8 & 73.6 \\
\bottomrule
\end{tabular}}
\end{table}
As shown in \Cref{tab:pareto3way}, the Pareto-optimal models share a consistent profile, i.e., all use curated or large-scale training data (DFN-5B or WebLI), SigLIP-based models or large CLIP backbones, and small patches and/or high resolution.
None of these models is the largest in the population (the biggest has 1.87B parameters vs.\ the 3.63B maximum), reinforcing that scale is not the path to the Pareto frontier.

Furthermore, as reported in \Cref{tab:consistent}, only 6 models appear in the top-20 on \emph{both} bias benchmarks simultaneously. These ``safe choices'' share the same characteristics as the Pareto-front models.
\begin{table}[t]
\centering
\caption{Models in the top-20 on \textbf{both} bias benchmarks. IN reports the ImageNet top-1 accuracy, and CA and UC the CelebA and UrbanCars WGA.}
\label{tab:consistent}
\small\resizebox{\linewidth}{!}{%

\begin{tabular}{lccc}
\toprule
\textbf{Model} & \textbf{IN (\%)} & \textbf{CA (\%)} & \textbf{UC (\%)} \\
\midrule
ViT-SO400M-14-SigLIP-384\_webli   & 83.1 & 89.2 & 86.4 \\
ViT-H-14-quickgelu\_dfn5b      & 83.4 & 87.0 & 84.8 \\
ViT-B-16-SigLIP2\_webli    & 79.2 & 87.2 & 83.2 \\
ViT-SO400M-14-SigLIP-378\_webli   & 83.1 & 88.3 & 81.6 \\
ViT-SO400M-16-SigLIP-i18n-256\_webli  & 81.5 & 87.8 & 80.0 \\
ViT-bigG-14\_laion2b\_s39b\_b160k      & 80.1 & 87.6 & 78.4 \\
\bottomrule
\end{tabular}}
\end{table}

\section{Discussion and Limitations}\label{sec:limitations}

Our study is scoped to publicly available checkpoints aggregated in the OpenCLIP repository. While this excludes proprietary models, it reflects the realistic setting faced by most practitioners and researchers, who select from openly available checkpoints. 
A further limitation is that all evaluated models have at most 3.6B parameters; larger models were excluded due to computational constraints.

The conducted evaluation uses two bias-related benchmarks (CelebA and UrbanCars) that were deliberately chosen to differ in spurious correlation complexity, enabling a controlled progression from single- to multi-attribute bias. These benchmarks are well-established in the spurious correlation literature~\cite{sagawa2020distributionally, li2023whac} and have known group structures, which is a prerequisite for reliable worst-group evaluation. Extending the audit to additional domains (e.g., medical imaging) is a natural direction for future work.

Finally, all evaluations are conducted in the zero-shot setting, which is both the standard deployment mode for VLMs and the setting where pre-trained biases are most directly exposed to end users, with no task-specific correction stage. 
Given the sensitivity of zero-shot performance to prompt selection, we conducted additional experiments on CelebA and UrbanCars using the 80 prompt templates from the ImageNet protocol described in \Cref{sec:protocol}. Results follow the same overall scaling trends, i.e., the correlation between model scale and performance progressively weakens from ImageNet to CelebA to UrbanCars (\rhoVal{0.68} $\rightarrow$ \rhoVal{0.49} $\rightarrow$ \rhoVal{0.19}), whereas the correlation with training-data size remains comparatively stable (\rhoVal{0.53} $\rightarrow$ \rhoVal{0.52} $\rightarrow$ \rhoVal{0.49}).

\section{Conclusion}\label{sec:conclusion}

 We presented the first large-scale empirical study of spurious correlation robustness across \nummodels{} publicly available VLMs, systematically examining how model scale, training data, and architectural choices affect WGA under biases of increasing complexity. Our central finding is that while scaling reliably improves ImageNet accuracy ($+2.6\%$ on average), it yields no benefit for multi-attribute bias scenarios and, in controlled comparisons, actually \emph{reduces} UrbanCars WGA by $4.2\%$ on average. 
In contrast, training data properties emerge as the most reliable lever for robustness. Dataset size retains a meaningful correlation with WGA even under multi-attribute bias, and data curation produces gains of up to 25\% over size-matched unfiltered alternatives. Architectural choices related to token granularity also play a consistent role, though their effect is benchmark-dependent.
The Pareto-optimal models identified in our analysis share a consistent profile consisting of models trained on curated or large-scale datasets, fine-grained tokenization, and moderate rather than maximal scale. 
Overall, our results highlight the importance of data quality as a first-class design objective alongside scale, and indicate that bias benchmarks covering multiple simultaneous spurious attributes should be incorporated as standard evaluation tools.

\section*{Acknowledgments}
This research was supported by the EU Horizon Europe projects ELIAS (grant no. 101120237), ELLIOT (grant no. 101214398), and GRAIL (grant no. 101298421).

\bibliographystyle{ACM-Reference-Format}
\balance
\bibliography{references}
 
\end{document}